\documentclass[10pt,twocolumn,letterpaper]{article}

\usepackage{cvpr}
\usepackage{times}
\usepackage{epsfig}
\usepackage{graphicx}
\usepackage{amsmath}
\usepackage{amssymb}
\usepackage{stmaryrd}
\usepackage{subcaption}
\usepackage{booktabs}
\usepackage{caption}
\usepackage[breaklinks=true,bookmarks=false]{hyperref}

\cvprfinalcopy 

\def\cvprPaperID{****} 

\newcommand\blfootnote[1]{%
  \begingroup
  \renewcommand\thefootnote{}\footnote{#1}%
  \addtocounter{footnote}{-1}%
  \endgroup
}

\begin{document}

\title{Improving Semantic Segmentation via Video Propagation and Label Relaxation}

\author{Yi Zhu$^{1\ast}$ \quad Karan Sapra$^{2\ast}$ \quad Fitsum A. Reda$^{2}$ \quad Kevin J. Shih$^{2}$ \quad Shawn Newsam$^{1}$\\ \quad Andrew Tao$^{2}$ \quad Bryan Catanzaro$^{2}$\\
{$^{1}$University of California at Merced \quad $^{2}$Nvidia Corporation}\\
\tt\small $\{$yzhu25,snewsam$\}$@ucmerced.edu \quad $\{$ksapra,freda,kshih,atao,bcatanzaro$\}$@nvidia.com}

\maketitle

\begin{abstract}
Semantic segmentation requires large amounts of pixel-wise annotations to learn accurate models. 
In this paper, we present a video prediction-based methodology to scale up training sets by synthesizing new training samples in order to improve the accuracy of semantic segmentation networks. 
We exploit video prediction models' ability to predict future frames in order to also predict future labels. A joint propagation strategy is also proposed to alleviate mis-alignments in synthesized samples. We demonstrate that training segmentation models on datasets augmented by the synthesized samples leads to significant improvements in accuracy.
Furthermore, we introduce a novel boundary label relaxation technique that makes training robust to annotation noise and propagation artifacts along object boundaries.
Our proposed methods achieve state-of-the-art mIoUs of $83.5\%$ on Cityscapes and $82.9\%$ on CamVid. Our single model, without model ensembles, achieves $72.8\%$ mIoU on the KITTI semantic segmentation test set, which surpasses the winning entry of the ROB challenge 2018. Our code and videos can be found at \url{https://nv-adlr.github.io/publication/2018-Segmentation}.\blfootnote{$\ast$ indicates equal contribution.}
\end{abstract}

\section{Introduction}
\label{sec:intro}
Semantic segmentation is the task of dense per pixel predictions of semantic labels. Large improvements in model accuracy have been made in recent literature \cite{Zhao2017pspnet,Chen2018deeplabv3plus,Bulo2018inplaceABN}, in part due to the introduction of Convolutional Neural Networks (CNNs) for feature learning, the task's utility for self-driving cars, and the availability of larger and richer training datasets (\eg, Cityscapes~\cite{Cordts2016Cityscapes} and Mapillary Vista~\cite{Neuhold2017mapillaryVista}). While these models rely on large amounts of training data to achieve their full potential, the dense nature of semantic segmentation entails a prohibitively expensive dataset annotation process. For instance, annotating all pixels in a $1024\times2048$ Cityscapes image takes on average $1.5$ hours \cite{Cordts2016Cityscapes}. Annotation quality plays an important role for training better models. While coarsely annotating large contiguous regions can be performed quickly using annotation toolkits, finely labeling pixels along object boundaries is extremely challenging and often involves inherently ambiguous pixels. 

\begin{figure}[t]
\begin{center}
   \includegraphics[width=1.0\linewidth]{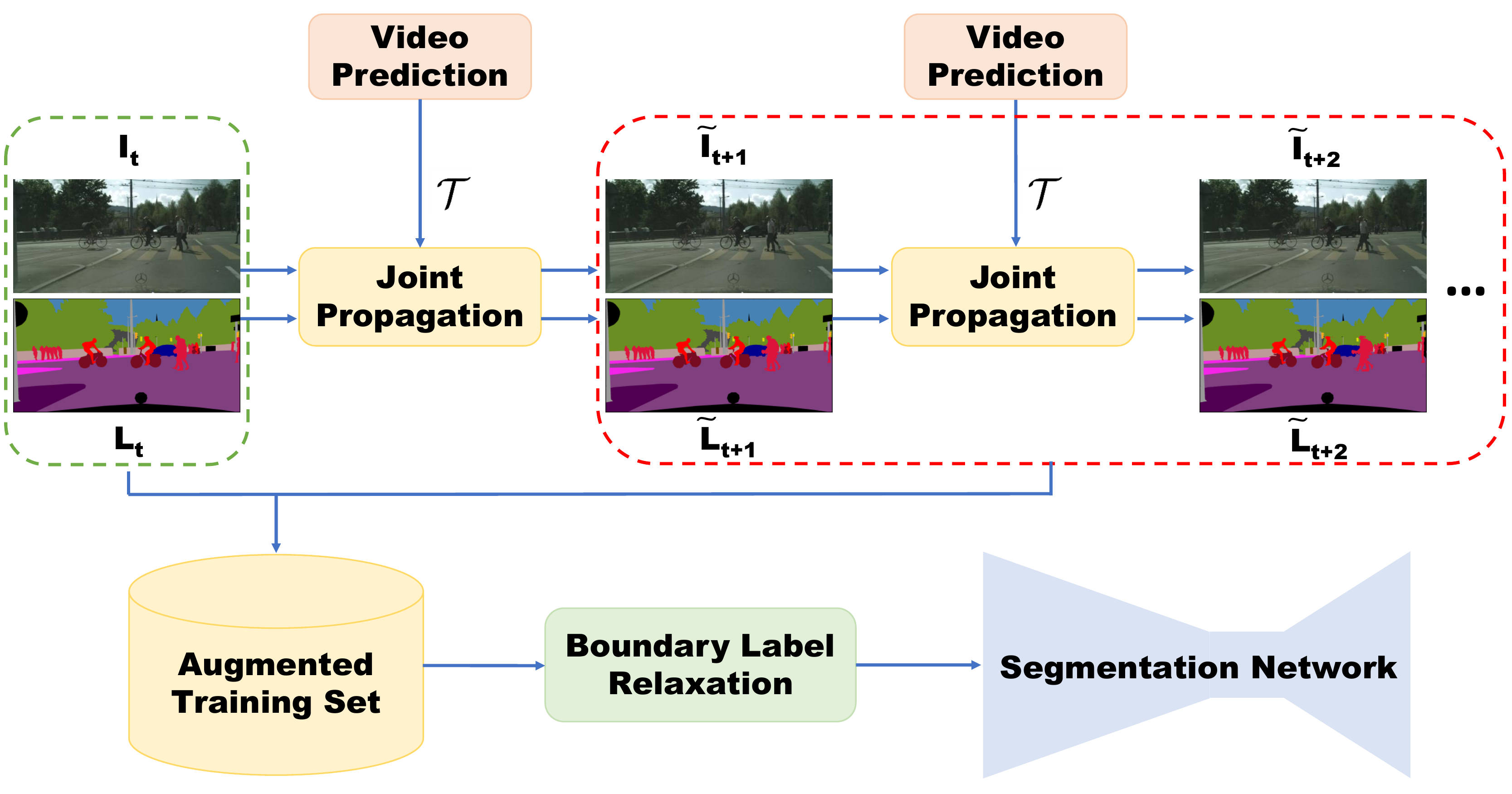}
   \vspace{-6ex}
\end{center}
   \caption{Framework overview. We propose joint image-label propagation to scale up training sets for robust semantic segmentation. The green dashed box includes manually labelled samples, and the red dashed box includes our propagated samples. $\mathcal{T}$ is the transformation function learned by the video prediction models to perform propagation. We also propose boundary label relaxation to mitigate label noise during training. Our framework can be used with most semantic segmentation and video prediction models.}
   \label{fig:first_page}
  \vspace{-2ex}
\end{figure}

Many alternatives have been proposed to augment training processes with additional data. For example, Cords \etal~\cite{Cordts2016Cityscapes} provided 20K coarsely annotated images to help train deep CNNs, an annotation cost effective alternative used by all top $10$ performers on the Cityscapes benchmark. Nevertheless, coarse labeling still takes, on average, $7$ minutes per image. 
An even cheaper way to obtain more labeled samples is to generate synthetic data \cite{SYNTHIA2016,Swami2018LSD,Hoffman2018cycada,Zlateski2018qualitySeg,Zhu2018ConservativeLoss}. However, model accuracy on the synthetic data often does not generalize to real data due to the domain gap between synthetic and real images. 
Luc \etal \cite{Luc2017futureSeg} use a state-of-the-art image segmentation method \cite{YuKoltun_dilate_ICLR2016} as a teacher to generate extra annotations for unlabelled images. However, their performance is bounded by the teacher method.
Another approach exploits the fact that many semantic segmentation datasets are based on continuous video frame sequences sparsely labeled at regular intervals. As such, several works \cite{Badrina2010labelProp,Budvytis2017augmentation,Mustikovela2016labelPropagation,gadde2017netwarp,Nilsson_2018_GRFP_CVPR} propose to use temporal consistency constraints, such as optical flow, to propagate ground truth labels from labeled to unlabeled frames. However, these methods all have different drawbacks which we will describe in Sec. \ref{sec:related}.

In this work, we propose to utilize video prediction models to efficiently create more training samples (image-label pairs) as shown in Fig. \ref{fig:first_page}.
Given a sequence of video frames having labels for only a subset of the frames in the sequence, we exploit the prediction models' ability to predict future frames in order to also predict future labels (new labels for unlabelled frames). Specifically, we propose leveraging such models in two ways. \textbf{1) Label Propagation (LP)}: We create new training samples by pairing a propagated label with the original future frame. \textbf{2) Joint image-label Propagation (JP)}: We create a new training sample by pairing a propagated label with the corresponding propagated image. In approach (2), it is of note that since both past labels and frames are jointly propagated using the same prediction model, the resulting image-label pair will have a higher degree of alignment. As we will show in later sections, we separately apply each approach for multiple future steps to scale up the training dataset. 

While great progress has been made in video prediction, it is still prone to producing unnatural distortions along object boundaries. For synthesized training examples, this means that the propagated labels along object boundaries should be trusted less than those within an object's interior. Here, we present a novel \textbf{boundary label relaxation} technique that can make training more robust to such errors. We demonstrate that by maximizing the likelihood of the \emph{union} of neighboring class labels along the boundary, the trained models not only achieve better accuracy, but are also able to benefit from longer-range propagation.

As we will show in our experiments, training segmentation models on datasets augmented by our synthesized samples leads to improvements on several popular datasets. Furthermore, by performing training with our proposed boundary label relaxation technique, we achieve even higher accuracy and training robustness, producing state-of-the-art results on the Cityscapes, CamVid, and KITTI semantic segmentation benchmarks. 
Our contributions are summarized below:

\begin{itemize}
	\item We propose to utilize video prediction models to propagate labels to immediate neighbor frames.
	\item We introduce joint image-label propagation to alleviate the mis-alignment problem. 
    \item We propose to relax one-hot label training by maximizing the likelihood of the union of class probabilities along boundary. This results in more accurate models and allows us to perform longer-range propagation.
    \item We compare our video prediction-based approach to standard optical flow-based ones in terms of segmentation performance.
\end{itemize}
     
\section{Related Work}
\label{sec:related}
Here, we discuss additional work related to ours, focusing mainly on the differences.

\noindent \textbf{Label propagation} 
There are two main approaches to propagating labels: patch matching \cite{Badrina2010labelProp,Budvytis2017augmentation} and optical flow \cite{Mustikovela2016labelPropagation,gadde2017netwarp,Nilsson_2018_GRFP_CVPR}. Patch matching-based methods, however, tend to be sensitive to patch size and threshold values, and, in some cases, they assume prior-knowledge of class statistics. Optical flow-based methods rely on very accurate optical flow estimation, which is difficult to achieve. Erroneous flow estimation can result in propagated labels that are misaligned with their corresponding frames. 

Our work falls in this line of research but has two major differences. First, we use motion vectors learned from video prediction models to perform propagation. The learned motion vectors can handle occlusion while also being class agnostic. Unlike optical flow estimation, video prediction models are typically trained through self-supervision. The second major difference is that we conduct joint image-label propagation to greatly reduce the mis-alignments.

\noindent \textbf{Boundary handling}
Some prior works \cite{Chen2016edgeSegmentation,Marmanis2018boundary} explicitly incorporate edge cues as constraints to handle boundary pixels. Although the idea is straightforward, this approach has at least two drawbacks. One is the potential error propagation from edge estimation and the other is fitting extremely hard boundary cases may lead to over-fitting at the test stage.
There is also literature focusing on structure modeling to obtain better boundary localization, such as affinity field \cite{Ke2018AAF}, random walk \cite{gberta_2017_CVPR}, relaxation labelling \cite{Vieux2012relaxation}, boundary neural fields \cite{gberta_2016_CVPR}, etc. However, none of these methods deals directly with boundary pixels but they instead attempt to model the interactions between segments along object boundaries. 
The work most similar to ours is
\cite{Kendall2017bayesUncertain} which proposes to incorporate uncertainty reasoning inside Bayesian frameworks. The authors enforce a Gaussian distribution over the logits to attenuate loss when uncertainty is large. Instead, we propose a modification to class label space that allows us to predict multiple classes at a boundary pixel. Experimental results demonstrate higher model accuracy and increased training robustness.

\begin{figure*}[t]
	\centering
	\includegraphics[trim={0 0 0 0},clip,width=1.0\linewidth]{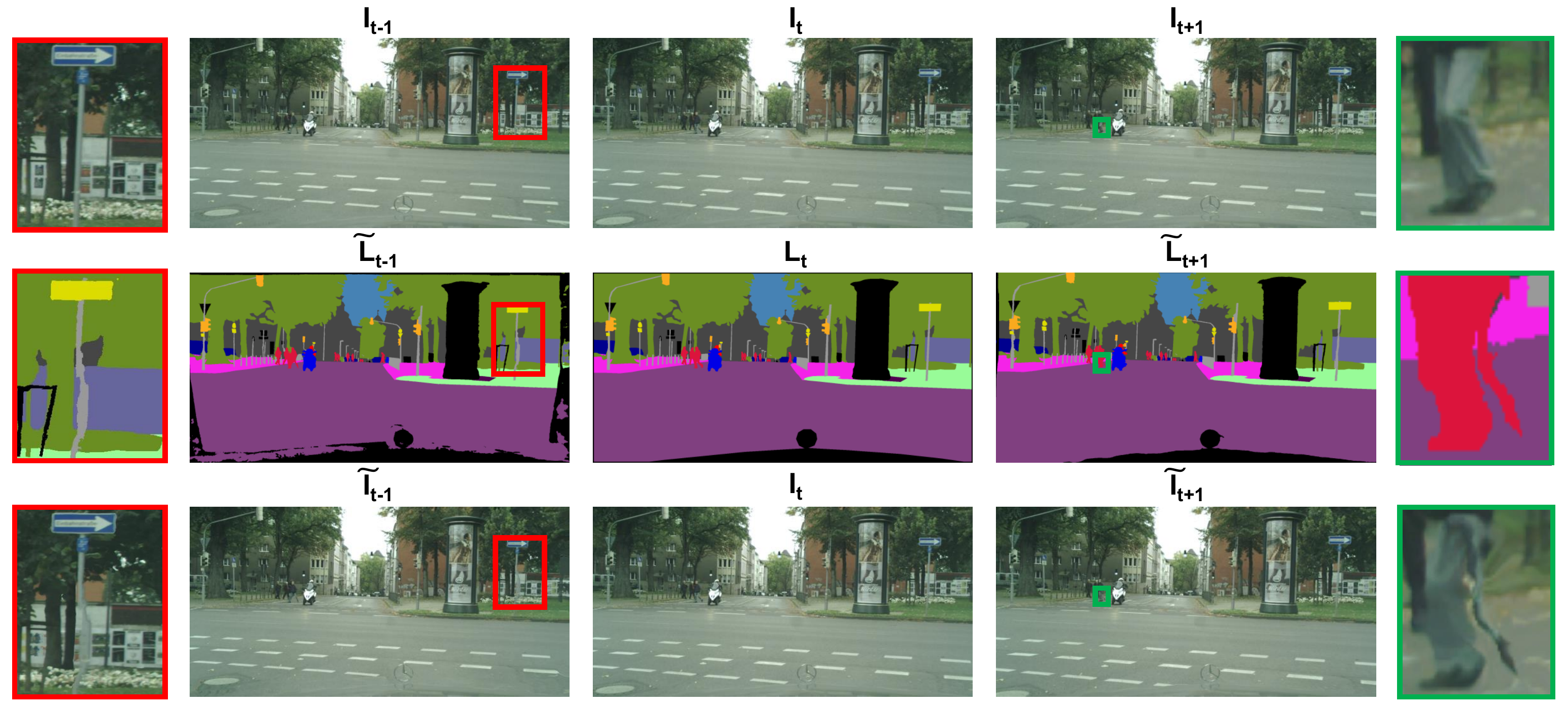}
	\vspace{-4ex}
	\caption{Motivation of joint image-label propagation. Row 1: original frames. Row 2: propagated labels. Row 3: propagated frames. The red and green boxes are two zoomed-in regions which demonstrate the mis-alignment problem. Note how the propagated frames align perfectly with propagated labels as compared to the original frames. The black areas in the labels represent a void class. (Image brightness has been adjusted for better visualization.)}
	\label{fig:s_warp}
	\vspace{-2ex}
\end{figure*}

\section{Methodology}
\label{sec:Methodology}
We present an approach for training data synthesis from sparsely annotated video frame sequences. Given an input video $\textbf{I} \in \Bbb R^{n \times W \times H}$ and semantic labels $\textbf{L} \in \Bbb R^{m \times W \times H}$, where $m \leq n$, we synthesize $k \times m$ new training samples (image-label pairs) using video prediction models, where $k$ is the length of propagation applied to each input image-label pair $(\textbf{I}_{i}, \textbf{L}_{i})$. We will first describe how we use video prediction models for label synthesis.

\subsection{Video Prediction}
\label{subsec:video_prediction}
Video prediction is the task of generating future frames from a sequence of past frames. 
It can be modeled as the process of direct pixel synthesis or learning to transform past pixels. 
In this work, we use a simple and yet effective vector-based approach \cite{Reda2018sdcnet} that predicts a motion vector $(u,v)$ to translate each pixel $(x,y)$ to its future coordinate. The predicted future frame ${\widetilde{\textbf{I}}_{t+1}}$ is given by,
\begin{equation} \label{eq:1}
{\widetilde{\textbf{I}}_{t+1}} = \mathcal{T}\Big(\mathcal{G}\big(\textbf{I}_{1:t}, \textbf{F}_{2:t}\big), \textbf{I}_{t}\Big) ,
\end{equation}
where $\mathcal{G}$ is a 3D CNN that predicts motion vectors $(u,v)$ conditioned on input frames $\textbf{I}_{1:t}$ and estimated optical flows $\textbf{F}_{i}$ between successive input frames $\textbf{I}_{i}$ and $\textbf{I}_{i-1}$. $\mathcal{T}$ is an operation that bilinearly samples from the most recent input $\textbf{I}_t$ using the predicted motion vectors $(u,v)$.

Note that the motion vectors predicted by $\mathcal{G}$ are not equivalent to optical flow vectors $\textbf{F}$. Optical flow vectors are undefined for pixels that are visible in the current frame but not visible in the previous frame. 
Thus, performing past frame sampling using optical flow vectors will duplicate foreground objects, create undefined holes or stretch image borders.  
The learned motion vectors, however, account for disocclusion and attempt to accurately predict future frames. We will demonstrate the advantage of learned motion vectors over optical flow in Sec. \ref{sec:experiments}. 

In this work, we propose to \emph{reuse} the predicted motion vectors to also synthesize future labels $\widetilde{\textbf{L}}_{t+1}$. Specifically:
\begin{equation} \label{eq:2}
{\widetilde{\textbf{L}}_{t+1}} = \mathcal{T}\Big(\mathcal{G}\big(\textbf{I}_{1:t}, \textbf{F}_{2:t}\big), \textbf{L}_{t}\Big) ,
\end{equation}
where a sampling operation $\mathcal{T}$ is applied on a past label $\textbf{L}_{t}$. $\mathcal{G}$ in equation \ref{eq:2} is the same as in equation \ref{eq:1} and is pre-trained on the underlying video frame sequences for the task of accurately predicting future frames.

\subsection{Joint Image-Label Propagation}
\label{subsec:joint_propagation}
Standard label propagation techniques create new training samples by pairing a propagated label with the original future frame as $\big(\textbf{I}_{i+k}, \widetilde{\textbf{L}}_{i+k}\big)$, with $k$ being the propagation length. For regions where the frame-to-frame correspondence estimation is not accurate, we will encounter mis-alignment between $\textbf{I}_{i+k}$ and $\widetilde{\textbf{L}}_{i+k}$.
For example, as we see in Fig. \ref{fig:s_warp}, most regions in the propagated label (row 2) correlate well with the  corresponding original video frames (row 1). 
However, certain regions, like the pole (red) and the leg of the pedestrian (green), do not align with the original frames due to erroneous estimated motion vectors.  

To alleviate this mis-alignment issue, we propose a joint image-label propagation strategy; \ie, we jointly propagate both the video frame and the label. 
Specifically, we apply equation \ref{eq:2} to each input training sample $(\textbf{I}_{i}, \textbf{L}_{i})$ for $k$ future steps to create $k \times m$ new training samples by pairing a predicted frame with a predicted label as ($\widetilde{\textbf{I}}_{i+k}, \widetilde{\textbf{L}}_{i+k}$).
As we can see in Fig. \ref{fig:s_warp}, the propagated frames (row 3) correspond well to the propagated labels (row 2). The pole and the leg experience the same distortion. Since semantic segmentation is a dense per-pixel estimation problem, such good alignment is crucial for learning an accurate model.

Our joint propagation approach can be thought of as a special type of data augmentation because both the frame and label are synthesized by transforming a past frame and the corresponding label using the same learned transformation parameters $(u,v)$. It is an approach similar to standard data augmentation techniques, such as random rotation, random scale or random flip. However, joint propagation uses a more fundamental transformation which was trained for the task of accurate future frame prediction. 

In order to create more training samples, we also perform reversed frame prediction. We equivalently apply joint propagation to create additional $k \times m$ new training samples as $(\widetilde{\textbf{I}}_{i-k}, \widetilde{\textbf{L}}_{i-k})$. In total, we can scale the training dataset by a factor of $2k+1$. In our study, we set $k$ to be $\pm 1, \pm 2, \pm 3, \pm 4$ or $\pm 5$, where $+$ indicates a forward propagation, and $-$ a backward propagation.

We would like to point out that our proposed joint propagation has broader applications. It could also find application in datasets where \emph{both} the raw frames and the corresponding labels are scarce. This is different from label propagation alone for synthesizing new training samples for typical video datasets, for instance Cityscapes \cite{Cordts2016Cityscapes}, where raw video frames are abundant but only a subset of the frames have human annotated labels. 

\begin{figure}[t]
\begin{center}
   \includegraphics[width=1.0\linewidth]{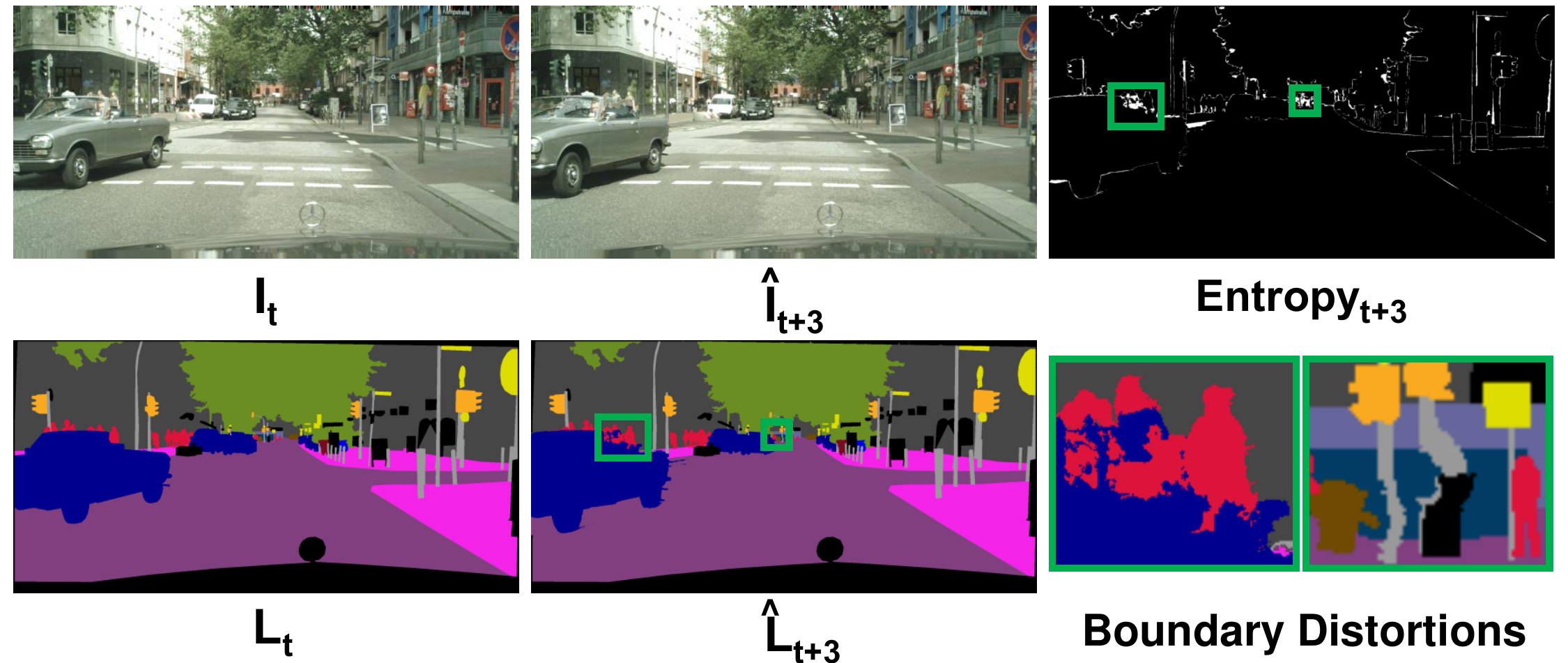}
   \vspace{-6ex}
\end{center}
   \caption{Motivation of boundary label relaxation. For the entropy image, the lighter pixel value, the larger the entropy. We find that object boundaries often have large entropy, due to ambiguous annotations or propagation distortions. The green boxes are zoomed-in figures showing such distortions.}
   \label{fig:rloss_sample}
   \vspace{-2ex}
\end{figure}

\subsection{Video Reconstruction}
\label{subsec:video_reconstruction}
Since, in our problem, we know the actual future frames, we can instead perform not just video prediction but video reconstruction to synthesize new training examples. 
More specifically, we can condition the prediction models on both the past and future frames to more accurately reconstruct ``future'' frames. The motivation behind this reformulation is that because future frames are observed by video reconstruction models, they are, in general, expected to produce better transformation parameters than video prediction models which only observe only past frames. 

Mathematically, a reconstructed future frame  $\hat{\textbf{I}}_{t+1}$ is given by,
\begin{equation} \label{eq:5}
\hat{\textbf{I}}_{t+1} = \mathcal{T}\Big(\mathcal{G}\big(\textbf{I}_{1:t+1}, \textbf{F}_{2:t+1}\big), \textbf{I}_{t}\Big) .
\end{equation}
In a similar way to equation \ref{eq:2}, we also apply $\mathcal{G}$ from equation \ref{eq:5} (which is learned for the task of accurate future frame reconstruction) to generate a future label $\hat{\textbf{L}}_{t+1}$. 

\begin{table}[t]
	\begin{center}
		\caption{Effectiveness of Mapillary	pre-training and class uniform sampling on both fine and coarse annotations. \label{tab:tricks}}
		\vspace{-2ex}
		\resizebox{0.8\columnwidth}{!}{%
			\begin{tabular}{  c  c  }
				\toprule
				Method	&     mIoU ($\%$)  \\
				\toprule		
				Baseline	   & $76.60$    \\
				+ Mapillary	Pre-training   & $78.32$ \\	
				+ Class Uniform Sampling (Fine + Coarse)  & $\mathbf{79.46}$ \\  
				\bottomrule
			\end{tabular}
		}
		\vspace{-4ex}
	\end{center}
\end{table} 

\subsection{Boundary Label Relaxation}
\label{subsec:relax}
Most of the hardest pixels to classify lie on the boundary between object classes \cite{Li2017DLC}. Specifically, it is difficult to classify the center pixel of a receptive field when potentially half or more of the input context could be from a different class. This problem is further compounded by the fact that the annotations are nowhere near pixel-perfect along the edges. 

We propose a modification to class label space, applied exclusively during training, that allows us to predict multiple classes at a boundary pixel. We define a boundary pixel as any pixel that has a differently labeled neighbor. Suppose we are classifying a pixel along the boundary of classes $A$ and $B$ for simplicity. Instead of maximizing the likelihood of the target label as provided by annotation, we propose to maximize the likelihood of $P(A \cup B)$. Because classes $A$ and $B$ are mutually exclusive, we aim to maximize the union of $A$ and $B$: 
\begin{equation}
P(A\cup B) = P(A) + P(B),
\end{equation}
where $P()$ is the softmax probability of each class. Specifically, let $\mathcal{N}$ be the set of classes within a 3$\times$3 window of a pixel. We define our loss as: 
\begin{equation}
    \mathcal{L}_{boundary} = -log \sum_{C \in \mathcal{N}}P(C).
\end{equation}
Note that for $|C| = 1$, this loss reduces to the standard one-hot label cross-entropy loss. 

One can see that the loss over the modified label space is minimized when $ \sum_{C \in \mathcal{N}}P(C) = 1$ without any constraints on the relative values of each class probability. We demonstrate that this relaxation not only makes our training robust to the aforementioned annotation errors, but also to distortions resulting from our joint propagation procedure. As can be seen in Fig. \ref{fig:rloss_sample}, the propagated label (three frames away from the ground truth) distorts along the moving car's boundary and the pole. Further, we can see how much the model is struggling with these pixels by visualizing the model's entropy over the class label . As the high entropy would suggest, the border pixel confusion contributes to a large amount of the training loss. In our experiments, we show that by relaxing the boundary labels, our training is more robust to accumulated propagation artifacts, allowing us to benefit from longer-range training data propagation.

\section{Experiments}
\label{sec:experiments}

In this section, we evaluate our proposed method on three widely adopted semantic segmentation datasets, including Cityscapes \cite{Cordts2016Cityscapes}, CamVid \cite{Brostow2008camvid} and KITTI \cite{Alhaija2018IJCV}. For all three datasets, we use the standard mean Intersection over Union (mIoU) metric to report segmentation accuracy.

\begin{table}[t]
	\begin{center}
		\caption{Comparison between (1) label propagation (LP) and joint propagation (JP); (2) video prediction (VPred) and video reconstruction (VRec). Using the proposed video reconstruction and joint propagation techniques, we improve over the baseline by $1.08\%$ mIoU ($79.46\% \shortrightarrow 80.54\%$).   \label{tab:reconstruction}}
		\vspace{-2ex}
		\resizebox{1.0\columnwidth}{!}{%
			\begin{tabular}{ c | c | c  c  c  c  c }
				\toprule
				&  0 & $\pm 1$ & $\pm 2$ & $\pm 3$ & $\pm 4$ & $\pm 5$ \\
				\toprule		
				VPred + LP &  $79.46$ & $79.79$ & $79.77$ &    $79.71$ &  $79.55$   &    $79.42$     \\
				VPred + JP & $79.46$ & $80.26$ &    $80.21$ &  $80.23$   &    $80.11$  & $80.04$     \\
				VRec + JP & $79.46$   & $\mathbf{80.54}$ &   $80.47$     &  $80.51$   &   $80.34$   & $80.18$     \\
				\bottomrule
			\end{tabular}
		}
		\vspace{-4ex}
	\end{center}
\end{table} 

\subsection{Implementation Details}
\label{subsec:train}
For the video prediction/reconstruction models, the training details are described in the supplementary materials. For semantic segmentation, we use an SGD optimizer and employ a polynomial learning rate policy \cite{Liu2017parsenet,Chen2018deeplabv2}, where the initial learning rate is multiplied by $(1 - \frac{epoch}{max\_epoch})^{power}$.
We set the initial learning rate to $0.002$ and power to $1.0$. Momentum and weight decay are set to $0.9$ and $0.0001$ respectively. We use synchronized batch normalization (batch statistics synchronized across each GPU) \cite{Zhao2017pspnet,Zhang_encnet_CVPR18} with a batch size of 16 distributed over 8 V100 GPUs. The number of training epochs is set to $180$ for Cityscapes, $120$ for Camvid and $90$ for KITTI. 
The crop size is $800$ for Cityscapes, $640$ for Camvid and $368$ for KITTI due to different image resolutions. 
For data augmentation, we randomly scale the input images (from 0.5 to 2.0), and apply horizontal flipping, Gaussian blur and color jittering during training. Our network architecture is based on DeepLabV3Plus \cite{Chen2018deeplabv3plus} with $output\_stride$ equal to $8$. 
For the network backbone, we use ResNeXt50 \cite{Xie2017ResNeXt} for the ablation studies, and WideResNet38 \cite{Wu2016WideOrDeep} for the final test-submissions. 
In addition, we adopt the following two effective strategies.

\noindent \textbf{Mapillary Pre-Training}
Instead of using ImageNet pre-trained weights for model initialization, we pre-train our model on Mapillary Vistas \cite{Neuhold2017mapillaryVista}. This dataset contains street-level scenes annotated for autonomous driving, which is close to Cityscapes. Furthermore, it has a larger training set (\ie, $18$K images) and more classes (\ie, $65$ classes). 

\noindent \textbf{Class Uniform Sampling}
We introduce a data sampling strategy similar to \cite{Bulo2018inplaceABN}. The idea is to make sure that all classes are approximately uniformly chosen during training. We first record the centroid of areas containing the class of interest. 
During training, we take half of the samples from the standard randomly cropped images and the other half from the centroids to make sure the training crops for all classes are approximately uniform per epoch. 
In this case, we are actually oversampling the underrepresented categories. 
For Cityscapes, we also utilize coarse annotations based on class uniform sampling. We compute the class centroids for all 20K samples, but we can choose which data to use.  
For example, classes such as fence, rider, train are underrepresented. Hence, we only augment these classes by providing extra coarse samples to balance the training. 

\subsection{Cityscapes}
\label{sec:cityscapes}

Cityscapes is a challenging dataset containing high quality pixel-level annotations for $5000$ images. The standard dataset split is $2975$, $500$, and $1525$ for the training, validation, and test sets respectively. There are also $20$K coarsely annotated images. All images are of size 1024$\times$2048. Cityscapes defines $19$ semantic labels containing both objects and stuff, and a void class for do-not-care regions. 
We perform several ablation studies below on the validation set to justify our framework design. 

\paragraph{Stronger Baseline}
First, we demonstrate the effectiveness of Mapillary pre-training and class uniform sampling. As shown in Table \ref{tab:tricks}, Mapillary pre-training is highly beneficial and improves mIoU by $1.72\%$ over the baseline ($76.60\% \shortrightarrow 78.32\%$). This makes sense because the Mapillary Vista dataset is close to Cityscape in terms of domain similarity, and thus provides better initialization than ImageNet.
We also show that class uniform sampling is an effective data sampling strategy to handle class imbalance problems. It brings an additional $1.14\%$ improvement ($78.32\% \shortrightarrow 79.46\%$). We use this recipe as our baseline.

\begin{figure}[t]
\begin{center}
   \includegraphics[width=1.0\linewidth]{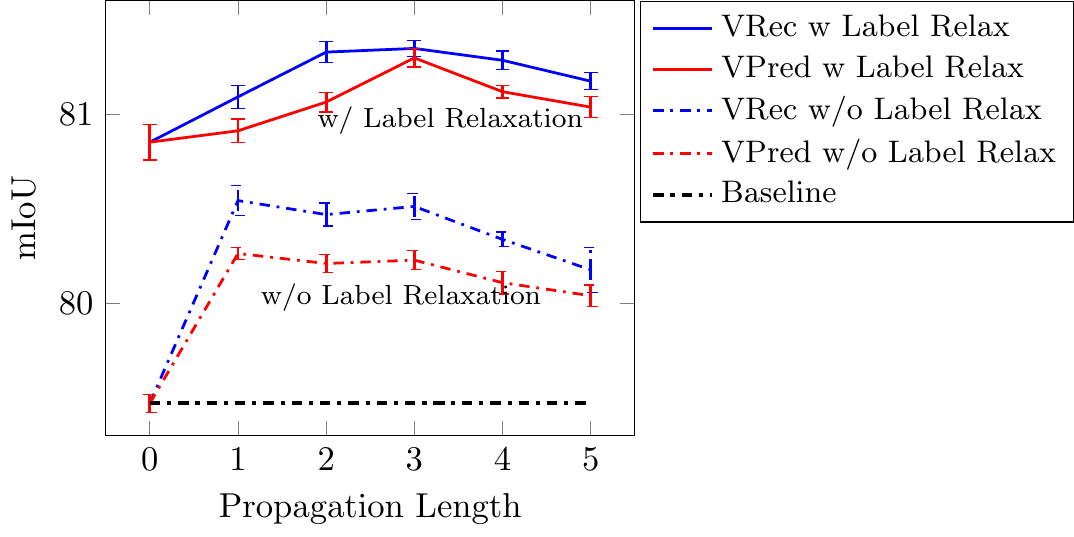}
   \vspace{-6ex}
\end{center}
   \caption{Boundary label relaxation leads to higher mIoU at all propagation lengths. The longer propagation, the bigger the gap between the solid (with label relaxation) and dashed (without relaxation) lines. The black dashed line represents our baseline ($79.46\%$). x-axis equal to 0 indicates no augmented samples are used. For each experiment, we perform three runs and report the mean and sample standard deviation as the error bar \cite{standard_error_1982}. }
   \label{fig:relax_help}
\end{figure}

\paragraph{Label Propagation versus Joint Propagation}
Next, we show the advantage of our proposed joint propagation over label propagation. For both settings, we use the motion vectors predicted by the video prediction model to perform propagation. The comparison results are shown in Table \ref{tab:reconstruction}.  
Column $0$ in Table \ref{tab:reconstruction} indicates the baseline ground-truth-only training (no augmentation with synthesized data).
Columns 1 to 5 indicate augmentation with sythesized data from timesteps $\pm k$, not including intermediate sythesized data from timesteps $<|k|$. For example, $\pm 3$ indicates we are using $+3$, $-3$ and the ground truth samples, but not $\pm 1$ and $\pm 2$. Note that we also tried the accumulated case, where $\pm 1$ and $\pm 2$ is included in the training set. However, we observed a slight performance drop. We suspect this is because the cumulative case significantly decreases the probability of sampling a hand-annotated training example within each epoch, ultimately placing too much weight on the synthesized ones and their imperfections. Comparisons between the non-accumulated and accumulated cases can be found in the supplementary materials. 

As we can see in Table~\ref{tab:reconstruction} (top two rows), joint propagation works better than label propagation at all propagation lengths. Both achieve highest mIoU for $\pm 1$, which is basically using information from just the previous and next frames. Joint propagation improves by $0.8\%$ mIoU over the baseline ($79.46\% \shortrightarrow 80.26\%$), while label propagation only improves by $0.33\%$ ($79.46\% \shortrightarrow 79.79\%$). This clearly demonstrates the usefulness of joint propagation. We believe this is because label noise from mis-alignment is outweighed by additional dataset diversity obtained from the augmented training samples. Hence, we adopt joint propagation in subsequent experiments. 

\begin{table*}[t]
	\begin{center}
		\caption{Per-class mIoU results on Cityscapes. Top: our ablation improvements on the validation set. Bottom: comparison with top-performing models on the test set. \label{tab:cs_sota}}
		\vspace{-2ex}
		\resizebox{2.1\columnwidth}{!}{%
			\begin{tabular}{  c | c | c  c  c  c  c c  c  c c  c  c c  c  c c  c  c c  c | c }
				\toprule
				Method	& split & road  & swalk & build. & wall & fence & pole & tlight & tsign & veg. & terrain & sky & person & rider & car & truck & bus & train & mcycle &  bicycle & mIoU   \\
				\hline
				\hline
				Baseline   &  val  &  $98.4$     &     $86.5$   &  $93.0$     &     $57.4$ &  $65.5$     &     $66.7$ &  $70.6$     &     $78.9$ &  $92.7$     &     $65.0$ &  $95.3$     &     $80.8$ &  $60.9$     &     $95.3$ &  $87.9$     &     $91.0$ &  $84.3$     &     $65.8$ &  $76.2$     &     $79.5$      \\
				+ VRec with JP  & val  &  $98.0$     &     $86.5$   &  $94.7$     &     $47.6$ &  $67.1$     &     $69.6$ &  $71.8$     &     $80.4$ &  $92.2$     &     $58.4$ &  $95.6$     &     $88.3$ &  $71.1$     &     $95.6$ &  $76.8$     &     $84.7$ &  $90.3$     &     $79.6$ &  $80.3$     &     $80.5$      \\
				+ Label Relaxation   & val  &  $98.5$     &     $87.4$   &  $93.5$     &     $64.2$ &  $66.1$     &     $69.3$ &  $74.2$     &     $81.5$ &  $92.9$     &     $64.6$ &  $95.6$     &     $83.5$ &  $66.5$     &     $95.7$ &  $87.7$     &     $91.9$ &  $85.7$     &     $70.1$ &  $78.8$     &     $81.4$      \\
				\hline
				\hline
				ResNet38 \cite{Wu2016WideOrDeep}   &  test  &  $98.7$     &     $86.9$   &  $93.3$     &     $60.4$ &  $62.9$     &     $67.6$ &  $75.0$     &     $78.7$ &  $93.7$     &     $73.7$ &  $95.5$     &     $86.8$ &  $71.1$     &     $96.1$ &  $75.2$     &     $87.6$ &  $81.9$     &     $69.8$ &  $76.7$     &     $80.6$      \\
				PSPNet  \cite{Zhao2017pspnet}   &  test     &  $98.7$     &     $86.9$   &  $93.5$     &     $58.4$ &  $63.7$     &     $67.7$ &  $76.1$     &     $80.5$ &  $93.6$     &     $72.2$ &  $95.3$     &     $86.8$ &  $71.9$     &     $96.2$ &  $77.7$     &     $91.5$ &  $83.6$     &     $70.8$ &  $77.5$     &     $81.2$        \\
				InPlaceABN \cite{Bulo2018inplaceABN} &  test     &  $98.4$     &     $85.0$   &  $93.6$     &     $61.7$ &  $63.9$     &     $67.7$ &  $77.4$     &     $80.8$ &  $93.7$     &     $71.9$ &  $95.6$     &     $86.7$ &  $72.8$     &     $95.7$ &  $79.9$     &     $93.1$ &  $89.7$     &     $72.6$ &  $78.2$     &     $82.0$       \\
				DeepLabV3+  \cite{Chen2018deeplabv3plus}&  test   &  $98.7$     &     $87.0$   &  $93.9$     &     $59.5$ &  $63.7$     &     $71.4$ &  $78.2$     &     $82.2$ &  $94.0$     &     $73.0$ &  $95.8$     &     $88.0$ &  $73.0$     &     $96.4$ &  $78.0$     &     $90.9$ &  $83.9$     &     $73.8$ &  $78.9$     &     $82.1$        \\
				DRN-CRL \cite{Zhuang2018DRN}  & test      &  $98.8$     &     $87.7$   &  $94.0$     &     $\mathbf{65.1}$ &  $64.2$     &     $70.1$ &  $77.4$     &     $81.6$ &  $93.9$     &     $73.5$ &  $95.8$     &     $88.0$ &  $74.9$     &     $96.5$ &  $\mathbf{80.8}$     &     $92.1$ &  $88.5$     &     $72.1$ &  $78.8$     &     $82.8$     \\
				Ours &  test &  $\mathbf{98.8}$     &     $\mathbf{87.8}$   &  $\mathbf{94.2}$     &     $64.1$ &  $\mathbf{65.0}$     &     $\mathbf{72.4}$ &  $\mathbf{79.0}$     &     $\mathbf{82.8}$ &  $\mathbf{94.2}$     &     $\mathbf{74.0}$ &  $\mathbf{96.1}$     &     $\mathbf{88.2}$ &  $\mathbf{75.4}$     &     $\mathbf{96.5}$ &  $78.8$     &     $\mathbf{94.0}$ &  $\mathbf{91.6}$     &     $\mathbf{73.8}$ &  $\mathbf{79.0}$     &     $\mathbf{83.5}$     \\
				\bottomrule
			\end{tabular}
		}
		\vspace{-4ex}
	\end{center}
\end{table*}

\paragraph{Video Prediction versus Video Reconstruction} Recall from Sec. \ref{subsec:video_prediction} that we have two methods for learning the motion vectors to generate new training samples through propagation: video prediction and video reconstruction. 
We experiment with both models in Table \ref{tab:reconstruction}.

As shown in Table \ref{tab:reconstruction} (bottom two rows), video reconstruction works better than video prediction at all propagation lengths, which agrees with our expectations. We also find that $\pm 1$ achieves the best result. Starting from $\pm 4$, the model accuracy starts to drop. This indicates that the quality of the augmented samples becomes lower as we propagate further. Compared to the baseline, we obtain an absolute improvement of $1.08\%$ ($79.46\% \shortrightarrow 80.54\%$). Hence, we use the motion vectors produced by the video reconstruction model in the following experiments.

\begin{figure}
\centering
\subcaptionbox{%
\begin{footnotesize}
    Top: MVs; Bottom: Flow
    \end{footnotesize}\label{fig:flow_vs_mv}
  }[0.48\linewidth]
  {\setlength{\tabcolsep}{0.1em} 
  \begin{tabular}{cc}
\includegraphics[width=.24\linewidth]{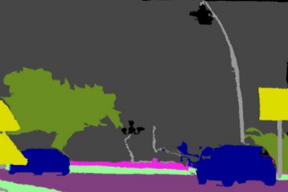} &
\includegraphics[width=.24\linewidth]{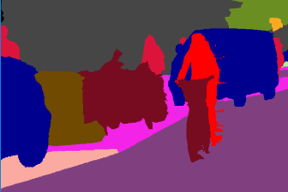}\\
\includegraphics[width=.24\linewidth]{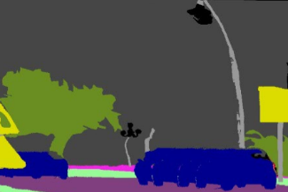}&
\includegraphics[width=.24\linewidth]{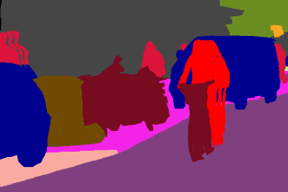} \\
\end{tabular}}
\subcaptionbox{%
 \begin{footnotesize}
    Propagation length performance \end{footnotesize}  \label{fig:flow_vs_mv_propagation}  
  }
  [0.5 \linewidth]{\includegraphics[width=.45\linewidth]{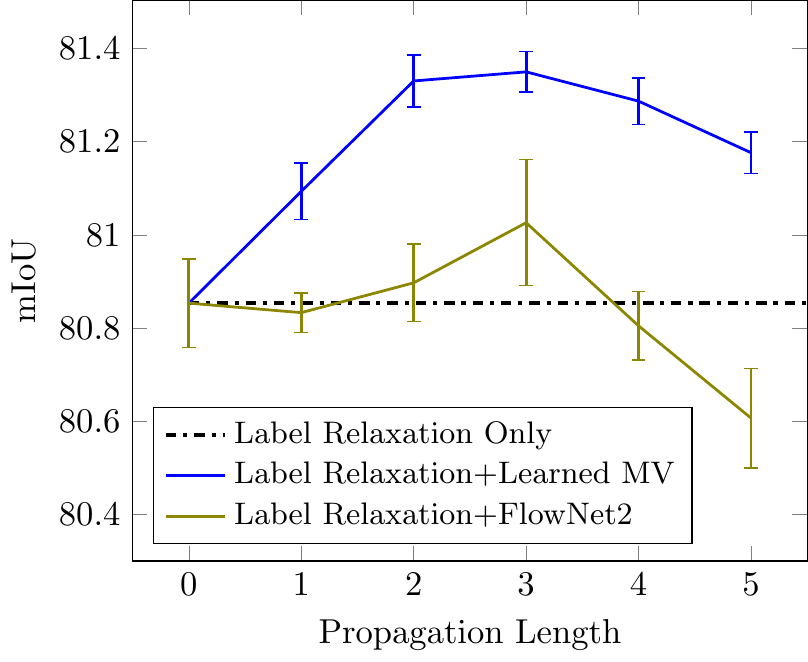}}
  \vspace{-4ex}
\caption{Our learned motion vectors from video reconstruction are better than optical flow (FlowNet2). \ref{fig:flow_vs_mv} Qualitative result. The learned motion vectors are better in terms of occlusion handling. \ref{fig:flow_vs_mv_propagation} Quantitative result. The learned motion vectors are better at all propagation lengths in terms of mIoU. 
}
\label{fig:sdc_vs_flow}
\vspace{-2ex}
\end{figure}

\paragraph{Effectiveness of Boundary Label Relaxation}
Theoretically, we can propagate the labels in an auto-regressive manner for as long as we want. The longer the propagation, the more diverse information we will get. However, due to abrupt scene changes and propagation artifacts, longer propagation will generate low quality labels as shown in Fig. \ref{fig:s_warp}. Here, we will demonstrate how the proposed boundary label relaxation technique can help to train a better model by utilizing longer propagated samples.  

We use boundary label relaxation on datasets created by video prediction (red) and video reconstruction (blue) in Fig. \ref{fig:relax_help}. 
As we can see, adopting boundary label relaxation leads to higher mIoU at all propagation lengths for both models. Take the video reconstruction model for example. Without label relaxation (dashed lines), the best performance is achieved at $\pm 1$. 
After incorporating relaxation (solid lines), the best performance is achieved at $\pm 3$ with an improvement of $0.81\%$ mIoU ($80.54\% \shortrightarrow 81.35\%$). The gap between the solid and dashed lines becomes larger as we propagate longer. The same trend can be observed for the video prediction models. This demonstrates that our boundary label relaxation is effective at handling border artifacts. It helps our model obtain more diverse information from $\pm 3$, and at the same time, reduces the impact of label noise brought by long propagation. Hence, we use boundary label relaxation for the rest of the experiments.

Note that even for no propagation (x-axis equal to 0) in Fig. \ref{fig:relax_help}, boundary label relaxation improves performance by a large margin ($79.46\% \shortrightarrow 80.85\%$). This indicates that our boundary label relaxation is versatile. Its use is not limited to reducing distortion artifacts in label propagation, but it can also be used in normal image segmentation tasks to handle ambiguous boundary labels. 

\paragraph{Learned Motion Vectors versus Optical Flow}

Here, we perform a comparison between the learned motion vectors from the video reconstruction model and optical flow, to show why optical flow is not preferred.
For optical flow, we use the state-of-the-art CNN flow estimator FlowNet2 \cite{Ilg2017flownet2} because it can generate sharp object boundaries and generalize well to both small and large motions. 

First, we show a qualitative comparison between the learned motion vectors and the FlowNet2 optical flow. As we can see in Fig. \ref{fig:flow_vs_mv}, FlowNet2 suffers from serious doubling effects caused by occlusion. For example, the dragging car (left) and the doubling rider (right). In contrast, our learned motion vectors can handle occlusion quite well. The propagated labels have only minor artifacts along the object borders which can be remedied by boundary label relaxation. Next, we show quantitative comparison between learned motion vectors and FlowNet2. As we can see in Fig. \ref{fig:flow_vs_mv_propagation}, the learned motion vectors (blue) perform significantly better than FlowNet2 (red) at all propagation lengths. As we propagate longer, the gap between them becomes larger, which indicates the low quality of the FlowNet2 augmented samples. Note that when the propagation length is $\pm 1, \pm 4$ and $\pm 5$, the performance of FlowNet2 is even lower than the baseline. 

\begin{figure}[t]
	\centering
	\includegraphics[trim={0 0 0 0},clip,width=1.0\linewidth]{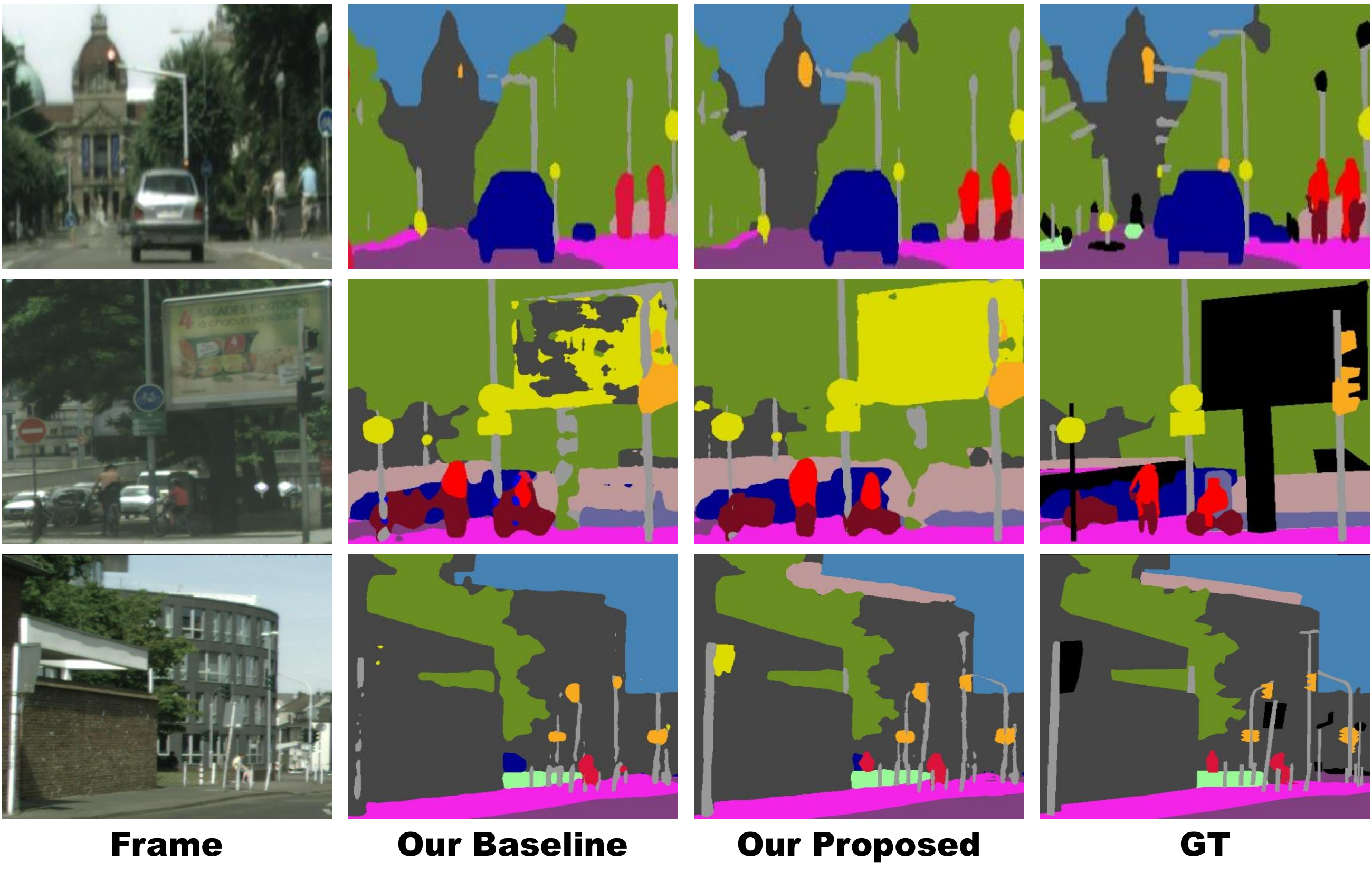}
	\vspace{-3ex}
	\caption{Visual comparisons on Cityscapes. The images are cropped for better visualization. We demonstrate our proposed techniques lead to more accurate segmentation than our baseline. Especially for thin and rare classes, like street light and bicycle (row 1), signs (row 2), person and poles (row 3). Our observation corresponds well to the class mIoU improvements in Table \ref{tab:cs_sota}. }
	\label{fig:cs_class_vis}
\end{figure}

\begin{figure}[t]
	\centering
	\includegraphics[trim={0 0 0 10},clip,width=1.0\linewidth]{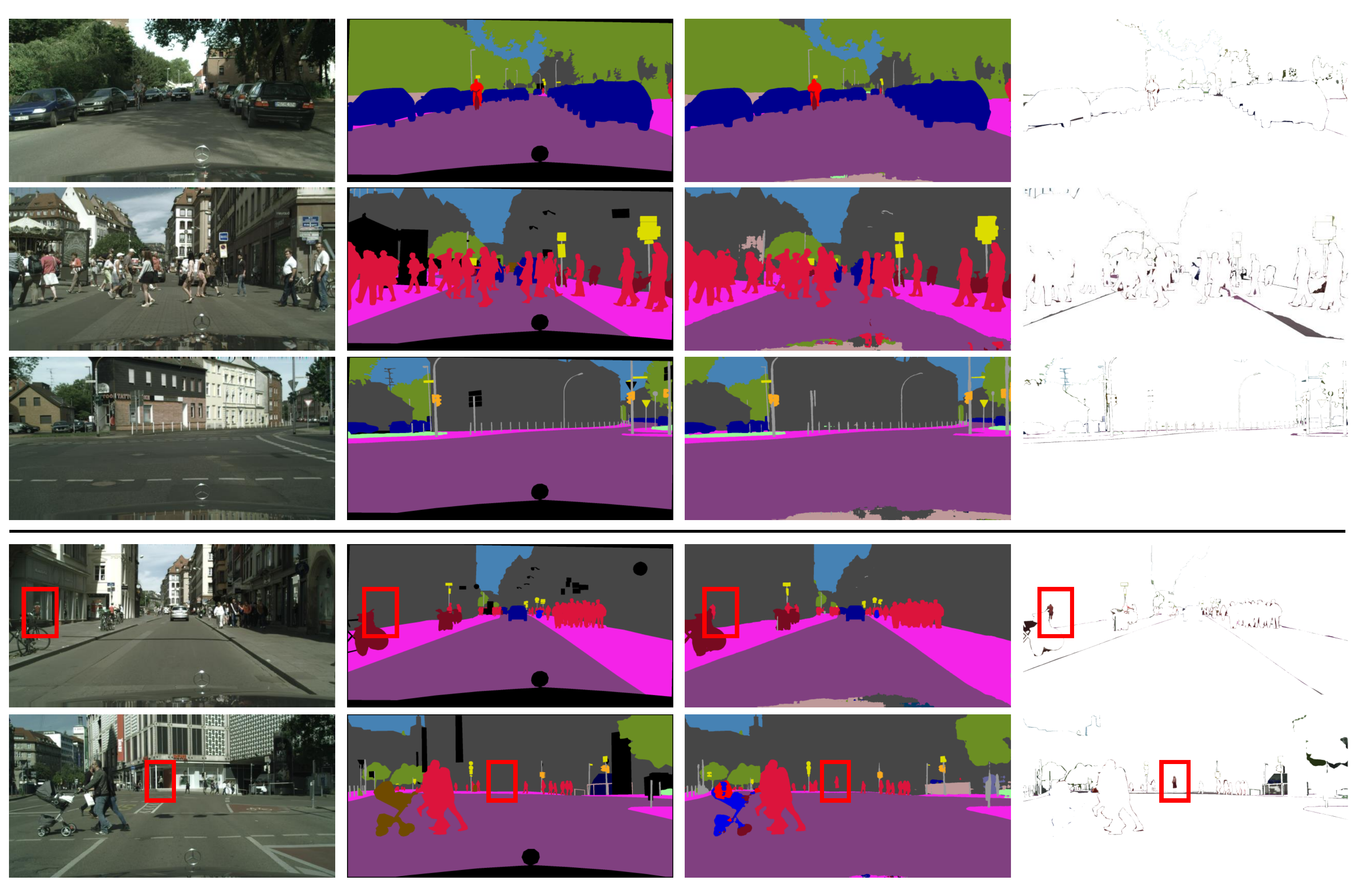}
	\vspace{-2ex}
	\caption{Visual examples on Cityscapes. From left to right: image, GT, prediction and their differences. We demonstrate that our model can handle situations with multiple cars (row 1), dense crowds (row 2) and thin objects (row 3). The bottom two rows show failure cases. We mis-classify a reflection in the mirror (row 4) and a model inside the building (row 5) as person (red boxes).}
	\label{fig:cs_kitti_vis}
\end{figure}

\begin{figure*}[t]
	\centering
	\includegraphics[trim={0 0 0 0},clip,width=1.0\linewidth]{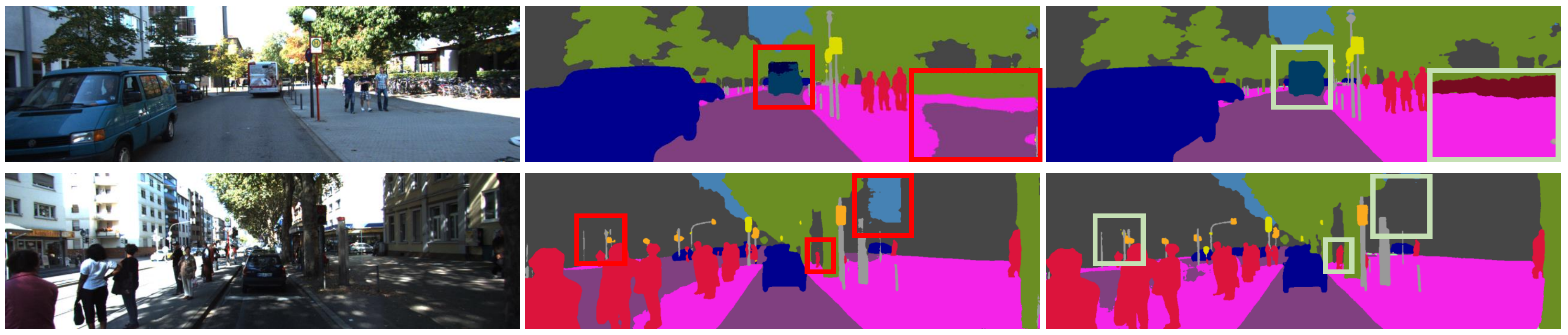}
	\vspace{-4ex}
	\caption{Visual comparison between our results and those of the winning entry \cite{Bulo2018inplaceABN} of ROB challenge 2018 on KITTI. From left to right: image, prediction from \cite{Bulo2018inplaceABN} and ours. Boxes indicate regions in which we perform better than \cite{Bulo2018inplaceABN}. Our model can predict semantic objects as a whole (bus), detect thin objects (poles and person) and distinguish confusing classes (sidewalk and road, building and sky).}
	\label{fig:kitti_map}
	\vspace{-2ex}
\end{figure*}

\paragraph{Comparison to State-of-the-Art}
As shown in Table \ref{tab:cs_sota} top, our proposed video reconstruction-based data synthesis together with joint propagation improves by $1.0\%$ mIoU over the baseline. Incorporating label relaxation brings another $0.9\%$ mIoU improvement. We observe that the largest improvements come from small/thin object classes, such as pole, street light/sign, person, rider and bicycle. 
This can be explained by the fact that our augmented samples result in more variation for these classes and helps with model generalization. We show several visual comparisons in Fig. \ref{fig:cs_class_vis}. 

For test submission, we train our model using the best recipe suggested above, and replace the network backbone with WideResNet38 \cite{Wu2016WideOrDeep}. We adopt a multi-scale strategy \cite{Zhao2017pspnet,Chen2018deeplabv3plus} to perform inference on multi-scaled (0.5, 1.0 and 2.0), left-right flipped and overlapping-tiled images, and compute the final class probabilities after averaging logits per inference.
More details can be found in the supplementary materials. 
As shown in Table \ref{tab:cs_sota} bottom, we achieve an mIoU of $83.5\%$, outperforming all prior methods. We get the highest IoU on $18$ out of the $20$ classes except for wall and truck.  
In addition, we show several visual examples in Fig. \ref{fig:cs_kitti_vis}. We demonstrate that our model can handle situations with multiple cars (row 1), dense crowds (row 2) and thin objects (row 3). 
We also show two interesting failure cases in Fig. \ref{fig:cs_kitti_vis}. Our model mis-classifies a reflection in the mirror (row 4) and a model inside the building (row 5) as person (red boxes). However, in terms of appearance without reasoning about context, our predictions are correct. 
More visual examples can be found in the supplementary materials.

\begin{table}[t]
	\begin{center}
		\caption{Results on the CamVid test set. Pre-train indicates the source dataset on which the model is trained.  \label{tab:camvid}}
		\vspace{-2ex}
		\resizebox{0.9\columnwidth}{!}{%
			\begin{tabular}{  c  c  c  c  c  }
				\toprule
				Method	&  Pre-train  & Encoder   & mIoU ($\%$)  \\
				\toprule
				SegNet \cite{Badrina2017segnet}	 & ImageNet & VGG16 &  $60.1$    \\
				RTA \cite{Huang_2018_ECCV} & ImageNet & VGG16 &$62.5$    \\
				Dilate8 \cite{YuKoltun_dilate_ICLR2016}	 & ImageNet & Dilate &  $65.3$    \\
				BiSeNet \cite{Yu_BiSeNet_2018eccv} 	 & ImageNet & ResNet18 &  $68.7$    \\
				PSPNet \cite{Zhao2017pspnet}	  & ImageNet& ResNet50 &  $69.1$    \\
				DenseDecoder \cite{Bilinski_2018_CVPR} & ImageNet	  & ResNeXt101 & $70.9$    \\
				VideoGCRF \cite{Chandra_2018_CVPR} 	& Cityscapes  & ResNet101  & $75.2$    \\ 
				\hline
				Ours (baseline)	 & Cityscapes & WideResNet38 & $79.8$    \\
				Ours 	 & Cityscapes & WideResNet38   & $\mathbf{81.7}$    \\
				\bottomrule
			\end{tabular}
		}
		\vspace{-4ex}
	\end{center}
\end{table}

\subsection{CamVid}
CamVid is one of the first datasets focusing on semantic segmentation for driving scenarios. It is composed of $701$ densely annotated images with size $720\times960$ from five video sequences. We follow the standard protocol proposed in \cite{Badrina2017segnet} to split the dataset into $367$ training, $101$ validation and $233$ test images. A total of $32$ classes are provided. However, most literature only focuses on $11$ due to the rare occurrence of the remaining classes.
To create the augmented samples, we directly use the video reconstruction model trained on Cityscapes without fine tuning on CamVid. The training strategy is similar to Cityscapes. 
We compare our method to recent literature in Table \ref{tab:camvid}. For fair comparison, we only report single-scale evaluation scores. 
As can be seen in Table \ref{tab:camvid}, we achieve an mIoU of $81.7\%$, outperforming all prior methods by a large margin. Furthermore, our multi-scale evaluation score is $82.9\%$. Per-class breakdown can be seen in the supplementary materials. 

One may argue that our encoder is more powerful than prior methods. To demonstrate the effectiveness of our proposed techniques, we perform training under the same settings without using the augmented samples and boundary label relaxation. The performance of this configuration on the test set is $79.8\%$, a significant IoU drop of $1.9\%$.   

\subsection{KITTI}
The KITTI Vision Benchmark Suite \cite{Geiger2012CVPR} was introduced in 2012 but updated with semantic segmentation ground truth \cite{Alhaija2018IJCV} in 2018. The data format and metrics conform with Cityscapes, but with a different image resolution of $375\times 1242$. The dataset consists of $200$ training and $200$ test images.
Since the dataset is quite small, we perform $10$-split cross validation fine-tuning on the $200$ training images. Eventually, we determine the best model in terms of mIoU on the whole training set because KITTI only allows one submission for each algorithm. For $200$ test images, we run multi-scale inference by averaging over $3$ scales ($1.5$, $2.0$ and $2.5$). We compare our method to recent literature in Table \ref{tab:kitti}. We achieve significantly better performance than prior methods on all four evaluation metrics. In terms of mIoU, we outperform previous state-of-the-art \cite{Bulo2018inplaceABN} by $3.3\%$. Note that \cite{Bulo2018inplaceABN} is the winning entry to Robust Vision Challenge 2018, which is achieved by an ensemble of five models, we only use one. We show two visual comparisons between ours and \cite{Bulo2018inplaceABN} in Fig. \ref{fig:kitti_map}.

\begin{table}[t]
	\begin{center}
		\caption{Results on KITTI test set.  \label{tab:kitti}}
		\vspace{-2ex}
		\resizebox{1.0\columnwidth}{!}{%
			\begin{tabular}{  c  c  c  c  c  }
				\toprule
				Method	&  IoU class &	iIoU class &	IoU category &	iIoU category  \\
				\toprule
				APMoE$\_$seg    \cite{kong2019pag} & $47.96$	& $17.86$	& $78.11$	& $49.17$ \\
				SegStereo \cite{yang2018segstereo}   & $59.10$	& $28.00$	& $81.31$	& $60.26$ \\
				AHiSS \cite{meletis2018AHiSS}	 & $61.24$	& $26.94$	& $81.54$	& $53.42$ \\
				LDN2 \cite{Ivan_ladderLDN_iccvw2017}	 & $63.51$	& $28.31$	& $85.34$	& $59.07$ \\
				MapillaryAI \cite{Bulo2018inplaceABN}	 & $69.56$	& $43.17$	& $86.52$	& $68.89$  \\
				\hline
				Ours 	 & $\mathbf{72.83}$ & $\mathbf{48.68}$ & $\mathbf{88.99}$  & $\mathbf{75.26}$    \\
				\bottomrule
			\end{tabular}
		}
		\vspace{-4ex}
	\end{center}
\end{table} 

\section{Conclusion}
\label{sec:conclusion}
We propose an effective video prediction-based data synthesis method to scale up training sets for semantic segmentation. We also introduce a joint propagation strategy to alleviate mis-alignments in synthesized samples. Furthermore, we present a novel boundary relaxation technique to mitigate label noise. The label relaxation strategy can also be used for human annotated labels and not just synthesized labels. 
We achieve state-of-the-art mIoUs of $83.5\%$ on Cityscapes, $82.9\%$ on CamVid, and $72.8\%$ on KITTI. The superior performance demonstrates the effectiveness of our proposed methods.

We hope our approach inspires other ways to perform data augmentation, such as GANs \cite{Liu2018GAN_SS}, to enable cheap dataset collection and achieve improved accuracy in target tasks. 
For future work, we would like to explore soft label relaxation using the learned kernels in \cite{Reda2018sdcnet} for better uncertainty reasoning. 
Our state-of-the-art implementation, will be made publicly available to the research community. 

\paragraph{Acknowledgements} We would like to thank Saad Godil, Matthieu Le, Ming-Yu Liu and Guilin Liu for suggestions and discussions.

{\small
\bibliographystyle{ieee}
\bibliography{egbib}
}

\newpage

\appendix 
\section*{Appendices}
\section{Implementation Details of Our Video Prediction/Reconstruction Models}
In this section, we first describe the network architecture of our video prediction model and then we illustrate the training details. The network architecture and training details of our video reconstruction model is similar, except the input is different. 

Recalling equation (1) from the main submission, the future frame ${\widetilde{\textbf{I}}_{t+1}}$ is given by,
\begin{equation*} 
{\widetilde{\textbf{I}}_{t+1}} = \mathcal{T}\Big(\mathcal{G}\big(\textbf{I}_{1:t}, \textbf{F}_{2:t}\big), \textbf{I}_{t}\Big) ,
\end{equation*}
where $\mathcal{G}$ is a general CNN that predicts the motion vectors $(u,v)$ conditioned on the input frames $\textbf{I}_{1:t}$ and the estimated optical flow $\textbf{F}_{i}$ between successive input frames $\textbf{I}_{i}$ and $\textbf{I}_{i-1}$. $\mathcal{T}$ is an operation that bilinearly samples from the most recent input $\textbf{I}_t$ using the predicted motion vectors $(u,v)$. 

In our implementation, we use the vector-based architecture as described in \cite{Reda2018sdcnet}. $\mathcal{G}$ is a fully convolutional U-net architecture, complete with an encoder and decoder and skip connections between encoder/decoder layers of the same output dimensions. 
Each of the $10$ encoder layers is composed of a convolution operation followed by a Leaky ReLU. The $6$ decoder layers are composed of a deconvolution operation followed by a Leaky ReLU. The output of the decoder is fed into one last convolutional layer to generate the motion vector predictions. The input to $\mathcal{G}$ is $\textbf{I}_{t-1}, \textbf{I}_{t}$ and $\textbf{F}_{t}$ (8 channels), and the output is the predicted 2-channel motion vectors that can best warp $\textbf{I}_{t}$ to $\textbf{I}_{t+1}$. For the video reconstruction model, we simply add $\textbf{I}_{t+1}$ and $\textbf{F}_{t+1}$ to the input, and change the number of channels in the first convolutional layer to $13$ instead of $8$.

We train our video prediction model using frames extracted from short sequences in the Cityscapes dataset. We use the Adam optimizer with $\beta_{1}=0.9$, $\beta_{2}=0.999$, and a weight decay of $1\times10^{-4}$. The frames are randomly cropped to $256\times256$ with no extra data augmentation. We set the batch size to 128 over 8 V100 GPUs. The initial learning rate is set to $1\times10^{-4}$ and the number of epochs is $400$. We refer interested readers to \cite{Reda2018sdcnet} for more details.  

\section{Non-Accumulated and Accumulated Comparison}

Recalling Sec. 4.1 from the main submission, we have two ways to augment the dataset. The first is the non-accumulated case, where we simply use synthesized data from timesteps $\pm k$, excluding intermediate synthesized data from timesteps $ < |k|$. For the accumulated case, we include all the synthesized data from timesteps $ \leq |k|$, which makes the augmented dataset $2k+1$ times larger than the original training set. 

We showed that we achieved the best performance at $\pm 3$, so we use $k=3$ here. We compare three configurations:
\begin{enumerate}
\item \textit{Baseline}: using the ground truth dataset only. 
\item \textit{Non-accumulated case}: using the union of the ground truth dataset and $\pm 3$; 
\item \textit{Accumulated case}: using the union of the ground truth dataset, $\pm 3$, $\pm 2$ and $\pm 1$.
\end{enumerate}
For these experiments, we use boundary label relaxation and joint propagation. We report segmentation accuracy on the Cityscapes validation set. 

\begin{table}[t]
	\begin{center}
		\caption{Accumulated and non-accumulated comparison. The numbers in brackets are the sample standard deviations.  \label{tab:accumulated}}
		\vspace{-1ex}
		\resizebox{1.0\columnwidth}{!}{%
			\begin{tabular}{  c  c  c  c }
				\toprule
				Method	&     Baseline  & Non-accumulated & Accumulated \\
				\toprule		
			    mIoU ($\%$) &	$80.85 \, (\pm 0.04)$	   & $81.35 \, (\pm 0.03)$  &   $81.12 \, (\pm 0.02)$\\ 
				\bottomrule
			\end{tabular}
		}
		\vspace{-2ex}
	\end{center}
\end{table} 

We have two observations from Table \ref{tab:accumulated}. First, using the augmented dataset always improves segmentation quality as quantified by mIoU. Second, the non-accumulated case performs better than the accumulated case. We suspect
this is because the cumulative case significantly decreases
the probability of sampling a hand-annotated training example within each epoch, ultimately placing too much weight on the synthesized ones and their imperfections. 

\section{Cityscapes}

\subsection{More Training Details}
We perform 3-split cross-validation to evaluate our algorithms, in terms of cities. The three validation splits are \{cv0: munster, lindau, frankfurt\}, \{cv1: darmstadt, dusseldorf, erfurt\} and \{cv2: monchengladbach, strasbourg, stuttgart\}. The rest cities will be in the training set, respectively. cv0 is the standard validation split. We found that models trained on cv2 split leads to higher performance on the test set, so we adopt cv2 split for our final test submission. 
Using our best model, we perform multiscale inference on the `stuttgart$\_$00' sequence and generate a demo video. The video is composed of both video frames and predicted semantic labels, with a $0.5$ alpha blending.

\subsection{Failure Cases}
\begin{figure*}[t]
	\centering
	\includegraphics[trim={0 0 0 0},clip,width=1.0\linewidth]{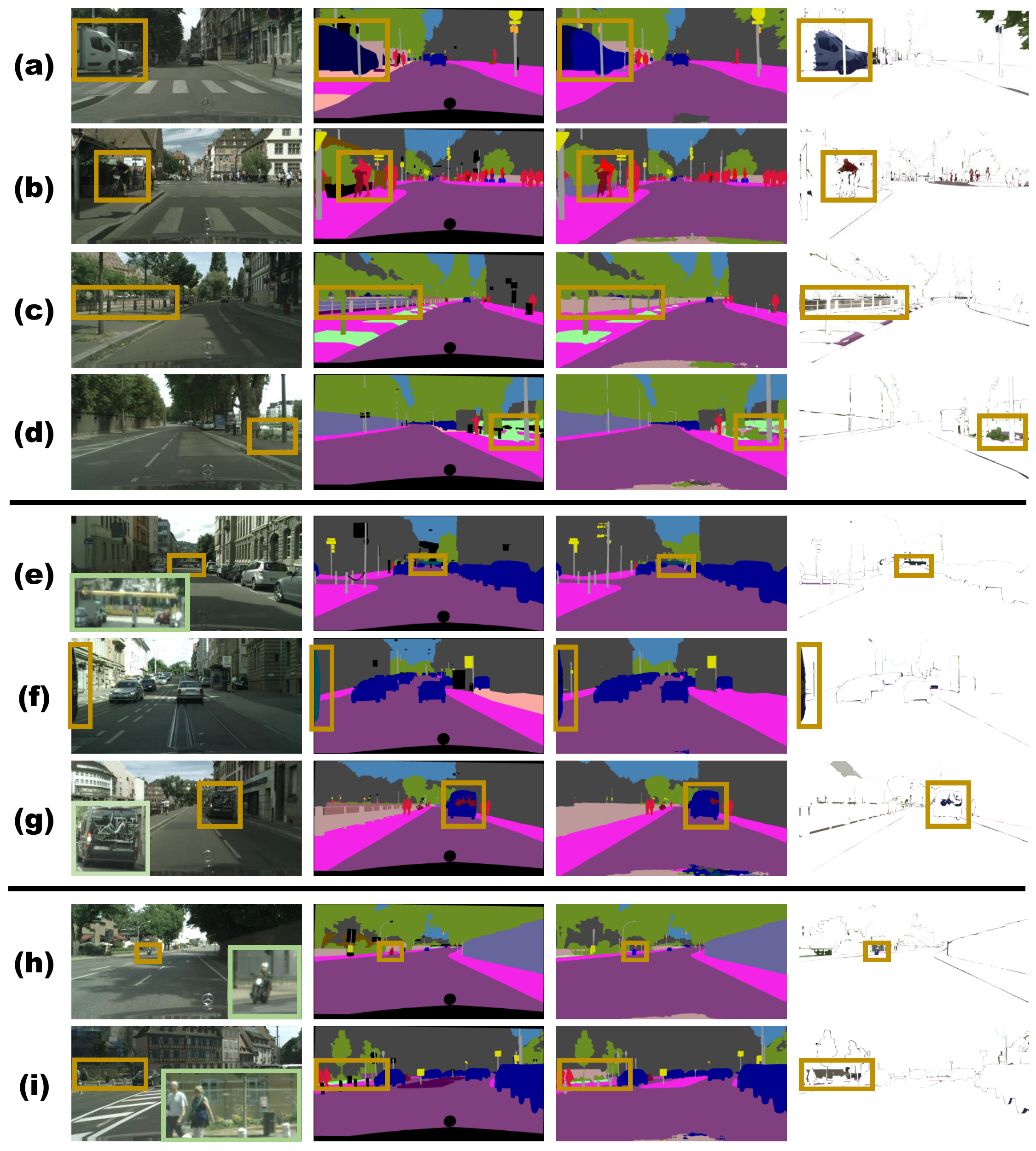}
	\caption{Failure cases (in yellow boxes). From left to right: image, ground truth, prediction and their difference. Green boxes are zoomed in regions for better visualization. Row (a) to (d) show class confusion problems. Our model has difficulty in segmenting: (a) car and truck. (b) person and rider. (c) wall and fence (d) terrain and vegetation. Row (e) to (f) show challenging cases when the object is far away, strongly occluded, or overlaps other objects. The last two rows show two training samples with wrong annotations: (h) mislabeled motorcycle to bicycle and (i) mislabeled fence to building.  }
	\label{fig:cs_fail}
\end{figure*}

We show several more failure cases in Fig. \ref{fig:cs_fail}.
First, we show four challenging scenarios of class confusion. From rows (a) to (d), our model has difficulty in segmenting: (a) car and truck. (b) person and rider. (c) wall and fence (d) terrain and vegetation.

Furthermore, we show three cases where it could be challenging even for a human to label. In Fig. \ref{fig:cs_fail} (e), it is very hard to tell whether it is a bus or train when the object is far away. In Fig. \ref{fig:cs_fail} (f), it is also hard to predict whether it is a car or bus under such strong occlusion (more than $95\%$ of the object is occluded). In Fig. \ref{fig:cs_fail} (g), there is a bicycle hanging on the back of a car. The model needs to know whether the bicycle is part of the car or a painting on the car, or whether they are two separate objects, in order to make the correct decision. 

Finally, we show two training samples where the annotation might be wrong. In Fig. \ref{fig:cs_fail} (h), the rider should be on a motorcycle, not a bicycle. In Fig. \ref{fig:cs_fail} (i), there should be a fence before the building. However, the whole region was labelled as building by a human annotator. In both cases, our model predicts the correct semantic labels. 

\subsection{More Synthesized Training Samples}
We show $15$ synthesized training samples in the demo video to give readers a better understanding. Each is a $11$-frame video clip, in which only the $5$th frame is the ground truth. The neighboring $10$ frames are generated using the video reconstruction model. We also show the comparison to using the video prediction model and FlowNet2 \cite{Ilg2017flownet2}. In general, the video reconstruction model gives us the best propagated frames/labels in terms of visualization. It also works the best in our experiments in terms of segmentation accuracy. 
Since the Cityscapes dataset is recorded at 17Hz \cite{Cordts2016Cityscapes}, the motion between frames is very large. Hence, propagation artifacts can be clearly observed, especially at the image borders. 

\section{CamVid}

\subsection{Class Breakdown}
We show the per-class mIoU results in Table \ref{tab:camvid_sota}. Our model has the highest mIoU on $8$ out of $11$ classes (all classes but tree, sky and sidewalk). This is expected because our synthesized training samples help more on classes with small/thin structures. Overall, our method significantly outperforms previous state-of-the-art by $7.7\%$ mIoU. 

\begin{table}[t]
	\begin{center}
		\caption{Per-class mIoU results on CamVid. Comparison with recent top-performing models on the test set. `SS' indicates single-scale inference, `MS' indicates multi-sclae inference. Our model achieves the highest mIoU on $8$ out of $11$ classes (all classes but tree, sky and sidewalk). This is expected because our synthesized training samples help more on classes with small/thin structures. \label{tab:camvid_sota}}
		\resizebox{1.0\columnwidth}{!}{%
			\begin{tabular}{  c | c   c c  c  c c  c  c c  c  c | c }
				\toprule
				Method	& Build. & Tree  & Sky & Car & Sign & Road & Pedes. & Fence & Pole & Swalk & Cyclist & mIoU   \\
				\toprule
				RTA \cite{Huang_2018_ECCV}   & $88.4$ & $\mathbf{89.3}$ &  $\mathbf{94.9}$ & $88.9$ & $48.7$ & $95.4$ & $73.0$ & $45.6$ & $41.4$ & $\mathbf{94.0}$ & $51.6$ & $62.5$ \\
				Dilate8 \cite{YuKoltun_dilate_ICLR2016}	 &      $82.6$ &  $76.2$     &     $89.0$   &  $84.0$     &     $46.9$ &  $92.2$    &     $56.3$ &  $35.8$     &     $23.4$ &  $75.3$     &     $55.5$      &     $65.3$     \\
				BiSeNet \cite{Yu_BiSeNet_2018eccv} &  $83.0$ & $75.8$ & $92.0$ & $83.7$ & $46.5$ & $94.6$ & $58.8$ & $53.6$ & $31.9$ & $81.4$ & $54.0$ & $68.7$ \\
				VideoGCRF \cite{Chandra_2018_CVPR}  &      $86.1$ &  $78.3$     &     $91.2$   &  $92.2$     &     $63.7$ &  $96.4$    &     $67.3$ &  $63.0$     &     $34.4$ &  $87.8$     &     $66.4$      &     $75.2$     \\
				\hline
				Ours (SS)  &      $90.9$ &  $82.9$     &     $92.8$   &  $\mathbf{94.2}$     &     $69.9$ &  $97.7$    &     $76.2$ &  $74.7$     &     $51.0$ &  $91.1$     &     $78.0$      &     $81.7$     \\
				Ours (MS) &      $\mathbf{91.2}$ &  $83.4$     &     $93.1$ &  $93.9$     &     $\mathbf{71.5}$ &  $\mathbf{97.7}$     &     $\mathbf{79.2}$ &  $\mathbf{76.8}$     &     $\mathbf{54.7}$ &  $91.3$     &     $\mathbf{79.7}$      &     $\mathbf{82.9}$     \\
				\bottomrule
			\end{tabular}
		}
		\vspace{-4ex}
	\end{center}
\end{table}

\subsection{More Synthesized Training Samples}
For CamVid, we show two demo videos of synthesized training samples. One is on the validation sequence `006E15', which is manually annotated every other frame. The other is on the training sequence `0001TP', which has manually annotated labels for every 30th frame. For `006E15', we do one step of forward propagation to generate a label for the unlabeled intermediate frame. For `0001TP', we do  $15$ steps of forward propagation and $14$ steps of backward propagation to label the $29$ unlabeled frames in between.  For both videos, the synthesized samples are generated using the video reconstruction model trained on Cityscapes, without fine-tuning on CamVid. This demonstrates the great generalization ability of our video reconstruction model.

\section{Demo Video}
We present all the video clips mentioned above at \url{https://nv-adlr.github.io/publication/2018-Segmentation}.

\end{document}